\documentclass{article}

\usepackage{arxiv}

\usepackage[utf8]{inputenc} 
\usepackage[T1]{fontenc}    
\usepackage{hyperref}       
\usepackage{url}           
\usepackage{booktabs}       
\usepackage{amsfonts}       
\usepackage{nicefrac}       
\usepackage{microtype}      
\usepackage{cleveref}
\usepackage{graphicx}
\usepackage{natbib}
\usepackage{doi}
\usepackage{tabularx}
\usepackage{multirow}

\title{JeFaPaTo - A joint toolbox for blinking analysis and facial features extraction}
\date{April 29, 2024}

\usepackage{authblk}

\newbox{\orcid}\sbox{\orcid}{\includegraphics[scale=0.06]{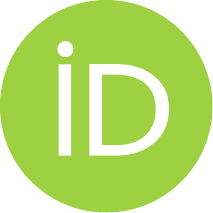}} 
\author[1]{%
	\href{https://orcid.org/000-0002-6879-552X}{\usebox{\orcid}\hspace{1mm}Tim Büchner\thanks{\texttt{tim.buechner@uni-jena.de}}}%
}
\author[1]{%
	\href{https://orcid.org/0000-0002-2294-3670}{\usebox{\orcid}\hspace{1mm}Oliver Mothes\thanks{\texttt{oliver.mothes@uni-jena.de}}}%
}
\author[2]{%
	\href{https://orcid.org/0000-0001-9671-0784}{\usebox{\orcid}\hspace{1mm}Orlando Guntinas-Lichius\thanks{\texttt{Orlando.Guntinas@med.uni-jena.de}}}%
}
\author[1]{%
	\href{https://orcid.org/0000-0002-3193-3300}{\usebox{\orcid}\hspace{1mm}Jochim Denzler\thanks{\texttt{joachim.denzler@uni-jena.de}}}%
}

\affil[1]{Computer Vision Group, Friedrich Schiller University Jena, 07743 Jena, Germany}
\affil[2]{Department of Otorhinolaryngology, Jena University Hospital, 07747 Jena, Germany}

\renewcommand{\undertitle}{A Preprint - Submitted to the Journal of Open Source Software}
\renewcommand{\shorttitle}{JeFaPaTo}

\hypersetup{
pdftitle={JeFaPaTo - A joint toolbox for blinking analysis and facial features extraction},
pdfsubject={Blinking Analysis},
pdfauthor={Tim Büchner, Oliver Mothes, Orlando Guntinas-Lichius, Joachim Denzler},
pdfkeywords={Python, Blinking, Facial Analysis, Blend Shapes, Facial Expressions},
}

\begin{document}
\maketitle

\section{Summary}
Analyzing facial features and expressions is a complex task in computer vision.
The human face is intricate, with significant shape, texture, and appearance variations.
In medical contexts, facial structures and movements that differ from the norm are particularly important to study and require precise analysis to understand the underlying conditions.
Given that solely the facial muscles, innervated by the facial nerve, are responsible for facial expressions, facial palsy can lead to severe impairments in facial movements~\citep{volkInitialSeverityMotor2017,louReviewAutomatedFacial2020}.

One affected area of interest is the subtle movements involved in blinking~\citep{vanderwerfBlinkRecoveryPatients2007,nuuttilaDiagnosticAccuracyGlabellar2021,vanderwerfEyelidMovementsBehavioral2003}.
It is an intricate spontaneous process that is not yet fully understood and needs high-resolution, time-specific analysis for detailed understanding~\citep{kwonHighspeedCameraCharacterization2013,cruzSpontaneousEyeblinkActivity2011}.
However, a significant challenge is that many computer vision techniques demand programming skills for automated extraction and analysis, making them less accessible to medical professionals who may not have these skills.
The Jena Facial Palsy Toolbox (JeFaPaTo) has been developed to bridge this gap.
It utilizes cutting-edge computer vision algorithms and offers a user-friendly interface for those without programming expertise.
This toolbox makes advanced facial analysis more accessible to medical experts, simplifying integration into their workflow.

This simple-to-use tool could enable medical professionals to quickly establish the blinking behavior of patients, providing valuable insights into their condition, especially in the context of facial palsy or Parkinson's disease~\citep{nuuttilaDiagnosticAccuracyGlabellar2021,vanderwerfBlinkRecoveryPatients2007}.
Due to facial nerve damage, the eye-closing process might be impaired and could lead to many undesirable side effects.
Hence, more than a simple distinction between open and closed eyes is required for a detailed analysis.
Factors such as duration, synchronicity, velocity, complete closure, the time between blinks, and frequency over time are highly relevant.
Such detailed analysis could help medical experts better understand the blinking process, its deviations, and possible treatments for better eye care.

\section{Statement of need}

To analyze the blinking behavior in detail, medical experts often use high-speed cameras to record the blinking process.
Existing tools modeling the eye state based on the Eye-Aspect-Ratio (EAR), such as~\cite{soukupovaRealTimeEyeBlink2016}, only classify the eye state as open or closed, requiring a labeled dataset for training a suitable classifier.
This approach neglects relevant information such as the blink intensity, duration, or partial blinks, which are crucial for a detailed analysis in a medical context.
Moreover, this simple classification approach does not factor in high temporal resolution video data, which is essential for a thorough analysis of the blinking process as most blinks are shorter than 100 ms.
We developed \texttt{JeFaPaTo} to go beyond the simple eye state classification and offer a method to extract complete blinking intervals for detailed analysis.
We aim to provide a custom tool that is easy for medical experts, abstracting the complexity of the underlying computer vision algorithms and high-temporal processing and enabling them to analyze blinking behavior without requiring programming skills.
An existing approach by \cite{kwonHighspeedCameraCharacterization2013} for high temporal videos uses only every frame 5 ms and requires manual measuring of the upper and lower eyelid margins.
Other methods require additional sensors such as electromyography (EMG) or magnetic search coils to measure the eyelid movement~\citep{vanderwerfBlinkRecoveryPatients2007,vanderwerfEyelidMovementsBehavioral2003}.
Such sensors necessitate additional human resources and are unsuitable for routine clinical analysis.
\texttt{JeFaPaTo} is a novel approach that combines the advantages of high temporal resolution video data~\citep{kwonHighspeedCameraCharacterization2013} and computer vision algorithms~\citep{soukupovaRealTimeEyeBlink2016} to analyze the blinking behavior.

\subsection{Overview of JeFaPaTo}

\texttt{JeFaPaTo} is a Python-based~\citep{pyqt6} program to support medical and psychological experts in analyzing blinking and facial features for high temporal resolution video data.
We follow a two-way approach to encourage programmers and non-programmers to use the tool.
On the one hand, we provide a programming interface for efficiently processing high-temporal resolution video data, automatic facial feature extraction, and specialized blinking analysis functions.
This interface is extendable, allowing the easy addition of new or existing facial feature-based processing functions (e.g., mouth movement analysis~\citep{hochreiterMachineLearningBasedDetectingEyelid2023} or MRD1/MRD2~\citep{chenSmartphoneBasedArtificialIntelligenceAssisted2021}.
On the other hand, we offer a graphical user interface (GUI) entirely written in Python to enable non-programmers to use the full analysis functions, visualize the results, and export the data for further analysis.
All functionalities of the programming interface are accessible through the GUI with additional input validations, making it easy for medical experts to use.
\texttt{JeFaPaTo} is designed to be extendable and transparent and is under joint development by computer vision and medical experts to ensure high usability and relevance for the target group.

\begin{figure}[t]
	\centering
	\includegraphics[width=1.0\textwidth]{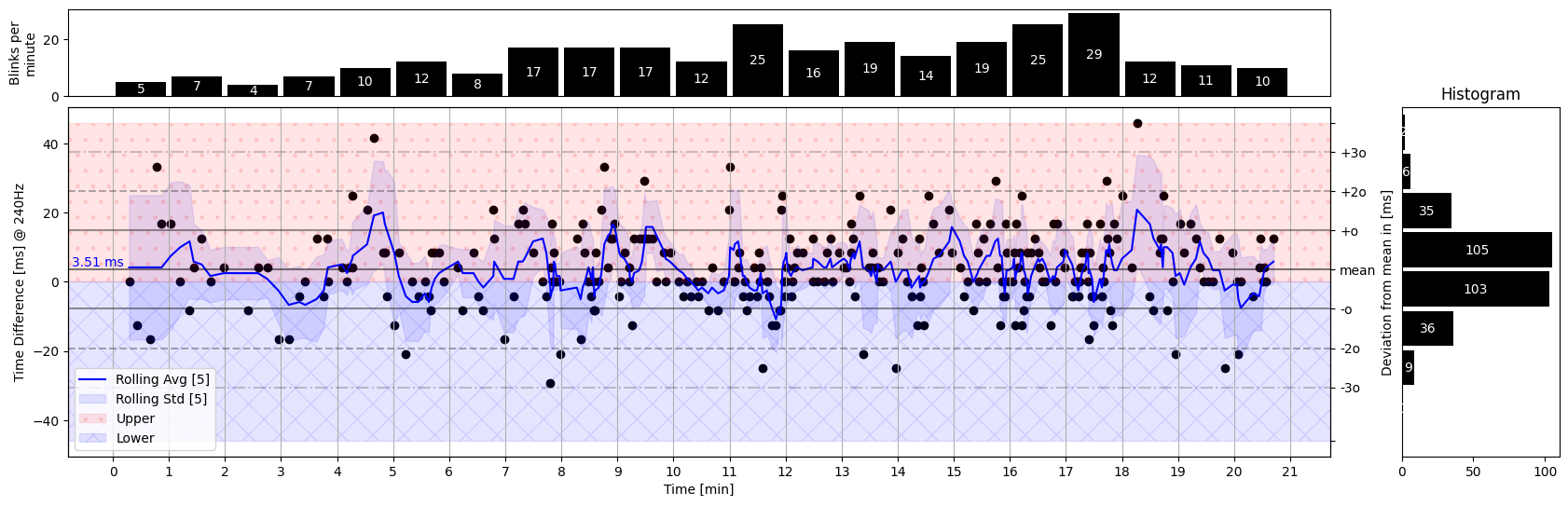}
	\label{fig:summary}
	\caption{
		The plot presents a visual summary of blinking patterns captured over 20 minutes, recorded at 240 frames per second (FPS).
		It illustrates the temporal variation in paired blinks, quantifies the blink rate as blinks per minute, and characterizes the distribution of the time discrepancy between left and right eye closures.
	}
\end{figure}

\texttt{JeFaPaTo} leverages the \emph{mediapipe} library~\citep{lugaresiMediaPipeFrameworkBuilding2019,kartynnikRealtimeFacialSurface2019a} to extract facial landmarks and blend shape features from video data at 60 FPS (on modern hardware).
With the landmarks, we compute the \emph{EAR} (Eye-Aspect-Ratio)~\cite{soukupovaRealTimeEyeBlink2016} for both eyes over the videos.
Additionally, \texttt{JeFaPaTo} detects blinks, matches the left and right eye, and computes medically relevant statistics.
Furthermore, a visual summary for the video is provided in the GUI, shown in \autoref{fig:summary}, and the data can be exported in various formats for further independent analysis.
The visual summary lets medical experts quickly get an overview of the blinking behavior.
As shown in \autoref{fig:summary}, the blinks per minute are shown as a histogram over time in the upper axis, and the delay between blinks is shown in the right axis.
The main plot comprises the scatter plot of the \emph{EAR} score for the left and right eye, and the dots indicate the detected blinks, with the rolling mean and standard deviation shown as a line plot.
This summary creates a compact overview by summarizing the blinking behavior throughout the video, enabling a quick individualized analysis for each video.

We leverage \texttt{PyQt6}~\citep{pyqt6,qt6} and \texttt{pyqtgraph}~\citep{pyqtgraph} to provide a GUI on any platform for easy usage.
To support and simplify the usage of \texttt{JeFaPaTo}, we provide a standalone executable for Windows, Linux, and MacOS.
\texttt{JeFaPaTo} is currently used in three medical studies to analyze the blinking process of healthy probands and patients with facial palsy and Parkinson's disease.

\section{Functionality and Usage}
\texttt{JeFaPaTo} was developed to support medical experts in extracting, analyzing, and studying blinking behavior.
Hence, the correct localization of facial landmarks is of high importance and the first step in the analysis process of each frame.
Once a user provides a video in the GUI, the tool performs an automatic face detection, and the user can adapt the bounding box if necessary.
Due to the usage of \texttt{mediapipe}~\citep{lugaresiMediaPipeFrameworkBuilding2019,kartynnikRealtimeFacialSurface2019a}, the tool can extract 468 facial landmarks and 52 blend shape features.
To describe the state of the eye, we use the Eye-Aspect-Ratio (EAR)~\citep{soukupovaRealTimeEyeBlink2016}, a standard measure for blinking behavior computed based on the 2D coordinates of the landmarks.
The ratio ranges between 0 and 1, where 0 indicates a fully closed eye and higher values indicate an open eye, whereas most people have an EAR score between 0.2 and 0.4.
This measure describes the ratio between the vertical and horizontal distance between the landmarks, resulting in a detailed motion approximation of the upper and lower eyelids.
Please note that all connotations for the left and right eye are based on the subject's viewing perspective.

\begin{figure}[t]
	\centering
	\includegraphics[width=0.7\textwidth]{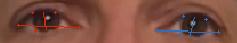}
	\label{fig:ear}
	\caption{
		Visualization of the Eye-Aspect-Ratio for the left (blue) and right (red) eye inside the face.
	}
\end{figure}

We denote this measure as \texttt{EAR-2D-6}, and the six facial landmarks are selected for both eyes, as shown in \autoref{fig:ear}. They are computed for each frame without any temporal smoothing.
As \texttt{mediapipe}~\citep{lugaresiMediaPipeFrameworkBuilding2019,kartynnikRealtimeFacialSurface2019a} belongs to the monocular depth reconstruction approaches for faces, each landmark contains an estimated depth value.
We offer the \texttt{EAR-3D-6} feature as an alternative, computed from 3D coordinates of the landmarks, to leverage this information to minimize the influence of head rotation.
However, the first experiments indicated that the 2D approach is sufficient to analyze blinking behavior.

\begin{figure}[t]
	\centering
	\includegraphics[width=0.8\textwidth]{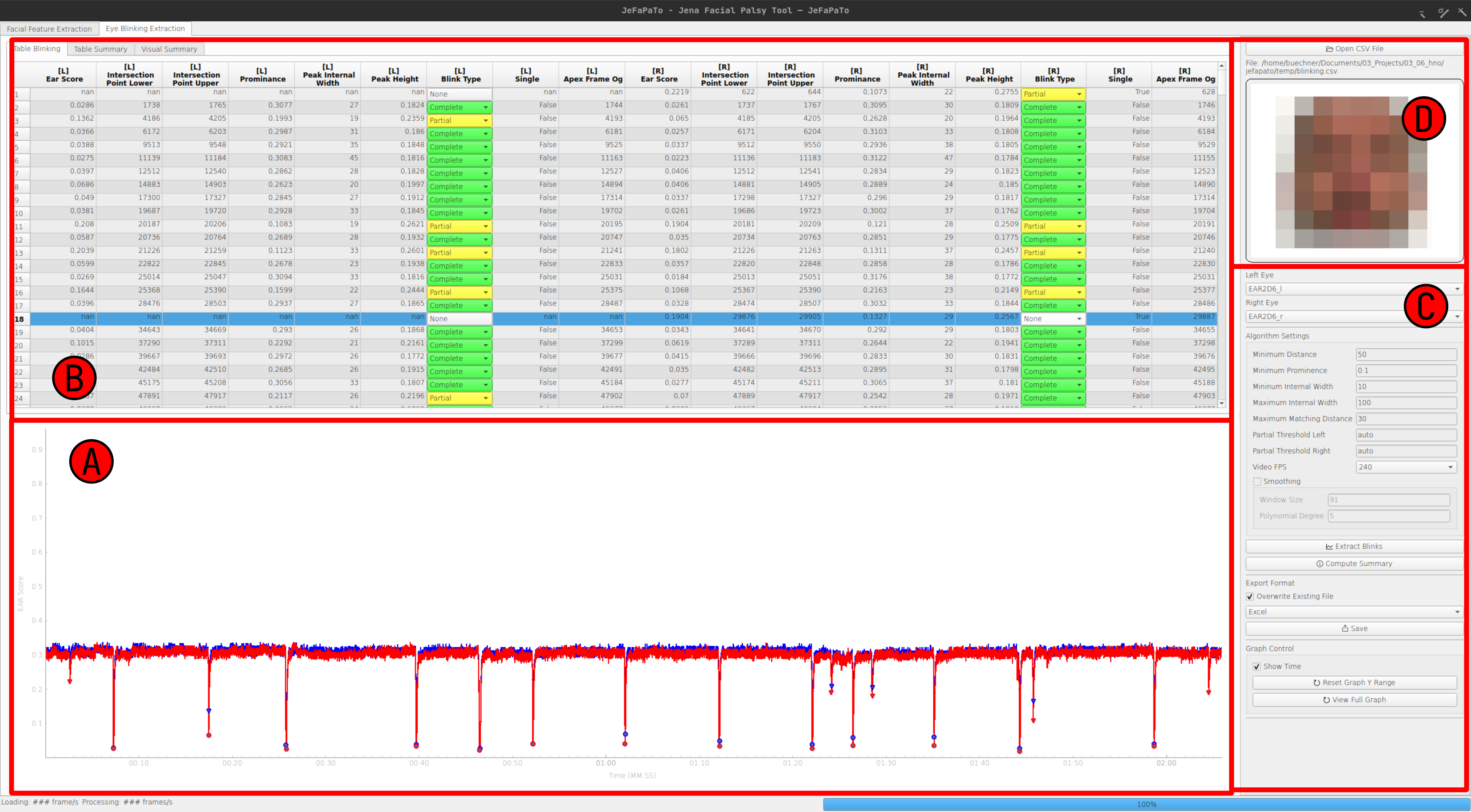}
	\label{fig:ui}
	\caption{
		The graphical user interface (GUI) designed for blinking analysis utilizes the Eye Aspect Ratio (EAR) metric. This interface comprises four primary components
	}
\end{figure}

\texttt{JeFaPaTo} optimizes io-read by utilizing several queues for loading and processing the video, assuring adequate RAM usage.
The processing pipeline extracts the landmarks and facial features, such as the `EAR` score for each frame, and includes a validity check ensuring that the eyes have been visible.
On completion, all values are stored in a CSV file for either external tools or for further processing \texttt{JeFaPaTo} to obtain insights into the blinking behavior of a person, shown in \autoref{fig:summary}.
The blinking detection and extraction employ the \texttt{scipy.signal.find\_peaks} algorithm~\citep{virtanenSciPyFundamentalAlgorithms2020}, and the time series can be smoothed if necessary.
We automatically match the left and right eye blinks based on the time of apex closure.
Additionally, we use the prominence of the blink to distinguish between `complete` and `partial` blinks based on a user-provided threshold (for each eye) or an automatic threshold computed using Otsu's method~\citep{otsu}.
The automatic threshold detection uses all extracted blinks for each eye individually.
Considering the personalized nature of blinking behavior, a graphical user interface (GUI) is provided, enabling experts to adjust the estimated blinking state as needed manually.
Additional functions are included in calculating blinking statistics: the blink rate (blinks per minute), the mean and standard deviation of the Eye Aspect Ratio (EAR) score, the inter-blink delay, and the blink amplitude.
A graphical user interface (GUI) for the \texttt{JeFaPaTo} codebase is provided, as depicted in \autoref{fig:ui}, to facilitate usage by individuals with limited programming expertise and to streamline data processing.

In \autoref{fig:ui}, we show the blinking analysis graphical user interface composed of four main areas.
We give a short overview of the functionality of each area to provide a better understanding of the tool's capabilities.
The A-Area is the visualization of the selected EAR time series for the left (drawn as a blue line) and right eye (drawn as a red line) over time.
Additionally, after successful blinking detection and extraction, the detected `complete` blinks (pupil not visible) are shown as dots, and `partial` blinks (pupil visible) as triangles.
If the user selects a blink in the table in the B-Area, the graph automatically highlights and zooms into the according area to allow a detailed analysis.

The B-Area contains the main table for the blinking extraction results, and the user can select the according blink to visualize the according period in the EAR plot.
The table contains the main properties of the blink: the EAR score at the blink apex, the prominence of the blink, the internal width in frames, the blink height, and the automatically detected blinking state (\texttt{none}, \texttt{partial}, \texttt{complete}).
If the user provides the original video, the user can drag and drop the video into the GUI into the D-Area, and the video will jump to the according frame to manually correct the blinking state.
The content of the table is used to compute the blinking statistics and the visual summary.
These statistics are also shown in the B-Area at different tabs, and the user can export the data as a CSV or Excel file for further analysis.

The C-Area is the control area, where the user can load the extracted EAR scores from a file and select the corresponding columns for the left and right eye (an automatic pre-selection is done).
The user can choose the parameters for the blinking extraction, such as the minimum prominence, distance between blinks, and the minimum blink width.
Additionally, users can define the decision threshold for estimating 'partial' blinks should the 'auto' mode prove inadequate.
Upon data extraction, corrections to the blinking state can be made directly within the table, following which the computation of blinking statistics and the generation of the visual summary can be initiated.

The D-Area displays the current video frame, given that the user supplies the original video.
While this feature is optional, it helps manually correct the blinking state when required.

\section{Extracted Medical Relevant Statistics}
We provided a set of relevant statistics for medical analysis of blinking behavior, which are valuable to healthcare experts, see \autoref{tab:statistics}.
The \texttt{JeFaPaTo} software is being developed in partnership with medical professionals to guarantee the included statistics are relevant.
Future updates may incorporate new statistics based on medical expert feedback.
A sample score file is available in the `examples/` directory within the repository, enabling users to evaluate the functionality of \texttt{JeFaPaTo} without recording a video.

\begin{table}[ht]
	\caption{Medical relevant statistics extracted by \texttt{JeFaPaTo} for blinking analysis.}
	\resizebox{\textwidth}{!}{%
		\begin{tabular}{lll}
			\toprule
			Statistic                        & Description                                                                      & Unit/Range      \\
			\midrule
			EAR\_Before\_Blink\_left\_avg    & The average left eye EAR score three seconds before the first blink              & $[0,1]$         \\
			EAR\_Before\_Blink\_right\_avg   & The average right eye EAR score three seconds before the first blink             & $[0,1]$         \\
			EAR\_left\_min                   & The minimum left eye EAR score of the time series                                & $[0,1]$         \\
			EAR\_right\_min                  & The minimum right eye EAR score of the time series                               & $[0,1]$         \\
			EAR\_left\_max                   & The maximum left eye EAR score of the time series                                & $[0,1]$         \\
			EAR\_right\_max                  & The maximum right eye EAR score of the time series                               & $[0,1]$         \\
			Partial\_Blink\_threshold\_left  & The threshold to distinguish `partial` or `complete` state                       & $[0,1]$         \\
			Partial\_Blink\_threshold\_right & The threshold to distinguish `partial` or `complete` state                       & $[0,1]$         \\
			Prominence\_min                  & The minimum prominence value of all blinks (left and right eye )                 & $[0,1]$         \\
			Prominence\_max                  & The maximum prominence value of all blinks (left and right eye )                 & $[0,1]$         \\
			Prominence\_avg                  & The average prominence value of all blinks (left and right eye )                 & $[0,1]$         \\
			Width\_min                       & The minimum width value of all blinks (left and right eye )                      & $[0,1]$         \\
			Width\_max                       & The maximum width value of all blinks (left and right eye )                      & $[0,1]$         \\
			Width\_avg                       & The average width value of all blinks (left and right eye )                      & $[0,1]$         \\
			Height\_min                      & The minimum height value of all blinks (left and right eye)                      & $[0,1]$         \\
			Height\_max                      & The maximum height value of all blinks (left and right eye)                      & $[0,1]$         \\
			Height\_avg                      & The average height value of all blinks (left and right eye)                      & $[0,1]$         \\
			Partial\_Blink\_Total\_left      & The amount of `partial` blinks for the left eye                                  & $\mathbb{N}$    \\
			Partial\_Blink\_Total\_right     & The amount of `partial` blinks for the right eye                                 & $\mathbb{N}$    \\
			Partial\_Frequency\_left\_bpm    & The frequency per minute of `partial` left blinks through out the video          & $\frac{1}{min}$ \\
			Partial\_Frequency\_right\_bpm   & The frequency per minute of `partial` right blinks through out the video         & $\frac{1}{min}$ \\
			Blink\_Length\_left\_ms\_avg     & The mean value of the blink length of the left eye                               & $ms$            \\
			Blink\_Length\_left\_ms\_std     & The standard deviation value of the  blink length of the left eye                & $ms$            \\
			Blink\_Length\_right\_ms\_avg    & The mean value of the blink length of the right eye                              & $ms$            \\
			Blink\_Length\_right\_ms\_std    & The standard deviation value of the  blink length of right left eye              & $ms$            \\
			Partial\_Blinks\_min[NN]\_left   & The amount of `partial` blinks in the left eye during minute $NN$ of the video   & $\mathbb{N}$    \\
			Partial\_Blinks\_min[NN]\_right  & The amount of `partial` blinks in the right eye during minute $NN$ of the video  & $\mathbb{N}$    \\
			Complete\_Blink\_Total\_left     & The amount of `complete` blinks in the left eye during minute $NN$ of the video  & $\mathbb{N}$    \\
			Complete\_Blink\_Total\_right    & The amount of `complete` blinks in the right eye during minute $NN$ of the video & $\mathbb{N}$    \\
			Complete\_Frequency\_left\_bpm   & The frequency per minute of `complete` left blinks through out the video         & $\frac{1}{min}$ \\
			Complete\_Frequency\_right\_bpm  & The frequency per minute of `complete` right blinks through out the video        & $\frac{1}{min}$ \\
			Complete\_Blinks\_min[NN]\_left  & The amount of `complete` blinks in the left eye during minute $NN$ of the video  & $\mathbb{N}$    \\
			Complete\_Blinks\_min[NN]\_left  & The amount of `complete` blinks in the right eye during minute $NN$ of the video & $\mathbb{N}$    \\
			\bottomrule                                                                                                                           \\
		\end{tabular}
	}
	\label{tab:statistics}
\end{table}

\section{Platform Support}
As \texttt{JeFaPaTo} is written in Python, it can be used on any platform that supports Python and the underlying libraries.
We recommend the usage of \texttt{anaconda}~\citep{anaconda} to create encapsulated Python environments to reduce the interference of already installed libraries and possible version mismatches.
The script \texttt{dev\_init.sh} automatically creates the custom environment with all dependencies with the \texttt{main.py} as the entry point for running the \texttt{JeFaPaTo}.
The user can also use the `requirements.txt` file to install the dependencies manually, even though we recommend creating a virtual environment at the very least.
As \texttt{JeFaPaTo} is designed to be used by medical experts, we provide a graphical user interface (GUI) to simplify usage during clinical studies and routine analysis.
We give each release a standalone executable for \texttt{Windows 11}, \texttt{Linux (Ubuntu 22.04)}, and \texttt{MacOS (version 13+ for Apple Silicon and Intel)}.
We offer a separate branch for \texttt{MacOS} version pre-13 (Intel), which does not contain blend shape extraction, to support older hardware.
The authors and medical partners conduct all user interface and experience tests on \texttt{Windows 11} and \texttt{MacOS 13+ (Apple Silicon)}.

\subsection{Extraction Parameter Recommendations}

The following parameters are recommended the blinking detection based on the current implementation of \texttt{JeFaPaTo}.
We list the settings for \emph{30 FPS} and \emph{240 FPS} videos and the time based parameters are measured in frames.
These settings can be adjusted in the GUI to adapt to the specific video data and the blinking behavior of the subject, if necessary.

\begin{table}[!ht]
	\caption{Recommended parameters for the blinking detection in \texttt{JeFaPaTo}.}
	\begin{tabular}{lll}
		\toprule
		Parameter                   & 30 FPS         & 240 FPS        \\
		\midrule
		Minimum Distance            & 10 Frames      & 50 Frames      \\
		Minimum Prominence          & 0.1 EAR Score  & 0.1 EAR Score  \\
		Minimum Internal Width      & 4 Frames       & 20 Frames      \\
		Maximum Internal Width      & 20 Frames      & 100 Frames     \\
		Maximum Matching Distance   & 15 Frames      & 30 Frames      \\
		Partial Threshold Left      & 0.18 EAR Score & 0.18 EAR Score \\
		Partial Threshold Right     & 0.18 EAR Score & 0.18 EAR Score \\
		Smoothing Window Size       & 7              & 7              \\
		Smoothing Polynomial Degree & 3              & 3              \\
		\bottomrule
	\end{tabular}
\end{table}

\subsection{Libraries}
We list the main libraries used in \texttt{JeFaPaTo} and their version used for the development.

\begin{table}[!ht]
	\begin{tabular}{llll}
		\toprule
		\textbf{Library}                & \textbf{Version}   & \textbf{Category} & \textbf{Usage}                                       \\
		\midrule
		\texttt{numpy}                  & \texttt{>=1.19}    & Processing        & Image matrices and time series                       \\
		\texttt{opencv-python-headless} & \texttt{>=4.5}     & Processing        & Image processing                                     \\
		\texttt{protobuf}               & \texttt{>=3.11,<4} & Processing        & Loading of precomputed models                        \\
		\texttt{psutil}                 & \texttt{~=5.8}     & Processing        & Computation of RAM requirements                      \\
		\texttt{mediapipe}              & \texttt{=0.10.8}   & Extraction        & Facial Feature Extraction                            \\
		\texttt{scipy}                  & \texttt{~=1.11}    & Extraction        & Extraction of blinks                                 \\
		\texttt{pandas}                 & \texttt{~=1.5}     & File Handling     & Loading and storing of files                         \\
		\texttt{openpyxl}               & \texttt{~=3.1}     & File Handling     & Support for Excel Files                              \\
		\texttt{matplotlib}             & \texttt{~=3.7}     & Plotting          & Creation of summary graphs                           \\
		\texttt{qtpy}                   & \texttt{~=2.3}     & GUI               & Simplified \texttt{Qt} interface for \texttt{Python} \\
		\texttt{qtawesome}              & \texttt{~=1.1}     & GUI               & FontAwesome Icons Interface for \texttt{Qt}          \\
		\texttt{PyQt6}                  & \texttt{~=6.5.3}   & GUI               & \texttt{Qt} Interface for \texttt{Python}            \\
		\texttt{PyQt6-Qt6}              & \texttt{~=6.5.3}   & GUI               & \texttt{Qt} Library                                  \\
		\texttt{pyqtgraph}              & \texttt{~=0.13}    & GUI               & Graph Visualization                                  \\
		\texttt{pyqtconfig}             & \texttt{~=0.9}     & GUI               & Saving and loading of user change settings           \\
		\texttt{pluggy}                 & \texttt{~=1.0}     & GUI               & Hook configuration between \texttt{JeFaPaTo} and GUI \\
		\texttt{structlog}              & \texttt{~=21.5}    & Logging           & Structured logging of information for development    \\
		\texttt{rich}                   & \texttt{~=12.0}    & Logging           & Colored logging                                      \\
		\texttt{plyer}                  & \texttt{~=2.1}     & Notifications     & Notification for the user for completed              \\
		\bottomrule
	\end{tabular}
\end{table}

\section{Ongoing Development}
\texttt{JeFaPaTo} finished the first stable release and will continue to be developed to support the analysis of facial features and expressions.
Given the potential of high temporal resolution video data to yield novel insights into facial movements, we aim to incorporate standard 2D measurement-based features into our analysis.
An issue frequently associated with facial palsy is synkinesis, characterized by involuntary facial muscle movements concurrent with voluntary movements of other facial muscles, such as the eye closing involuntarily when the patient smiles.
Hence, a joint analysis of the blinking pattern and mouth movement could help better understand the underlying processes.
The EAR is sensitive to head rotation.
Careful setting up the experiment can reduce the influence of head rotation, but it is not always possible.
To support the analysis of facial palsy patients, we plan to implement a 3D head pose estimation to correct the future EAR score for head rotation.

\section{Acknowledgements}
Supported by Deutsche Forschungsgemeinschaft (DFG - German Research Foundation) project 427899908 BRIDGING THE GAP: MIMICS AND MUSCLES (DE 735/15-1 and GU 463/12-1).
We acknowledge the helpful feedback for the graphical user interface development and quality-of-life requests from Lukas Schuhmann, Elisa Furche, Elisabeth Hentschel, and Yuxuan Xie.

\bibliographystyle{unsrtnat}
\bibliography{paper}

\begin{thebibliography}{18}
\providecommand{\natexlab}[1]{#1}
\providecommand{\url}[1]{\texttt{#1}}
\expandafter\ifx\csname urlstyle\endcsname\relax
  \providecommand{\doi}[1]{doi: #1}\else
  \providecommand{\doi}{doi: \begingroup \urlstyle{rm}\Url}\fi

\bibitem[Volk et~al.(2017)Volk, Granitzka, Kreysa, Klingner, and
  {Guntinas-Lichius}]{volkInitialSeverityMotor2017}
Gerd~Fabian Volk, Thordis Granitzka, Helene Kreysa, Carsten~M. Klingner, and
  Orlando {Guntinas-Lichius}.
\newblock Initial severity of motor and non-motor disabilities in patients with
  facial palsy: An assessment using patient-reported outcome measures.
\newblock \emph{European archives of oto-rhino-laryngology: official journal of
  the European Federation of Oto-Rhino-Laryngological Societies (EUFOS):
  affiliated with the German Society for Oto-Rhino-Laryngology - Head and Neck
  Surgery}, 274\penalty0 (1):\penalty0 45--52, January 2017.
\newblock ISSN 1434-4726.
\newblock \doi{10.1007/s00405-016-4018-1}.

\bibitem[Lou et~al.(2020)Lou, Yu, and Wang]{louReviewAutomatedFacial2020}
Jianwen Lou, Hui Yu, and Fei-Yue Wang.
\newblock A {{Review}} on {{Automated Facial Nerve Function Assessment From
  Visual Face Capture}}.
\newblock \emph{IEEE Transactions on Neural Systems and Rehabilitation
  Engineering}, 28\penalty0 (2):\penalty0 488--497, February 2020.
\newblock ISSN 1558-0210.
\newblock \doi{10.1109/TNSRE.2019.2961244}.

\bibitem[VanderWerf et~al.(2007)VanderWerf, Reits, Smit, and
  Metselaar]{vanderwerfBlinkRecoveryPatients2007}
Frans VanderWerf, Dik Reits, Albertine~Ellen Smit, and Mick Metselaar.
\newblock Blink {{Recovery}} in {{Patients}} with {{Bell}}'s {{Palsy}}: {{A
  Neurophysiological}} and {{Behavioral Longitudinal Study}}.
\newblock \emph{Investigative Ophthalmology \& Visual Science}, 48\penalty0
  (1):\penalty0 203--213, January 2007.
\newblock ISSN 1552-5783.
\newblock \doi{10.1167/iovs.06-0499}.

\bibitem[Nuuttila et~al.(2021)Nuuttila, Eklund, Joutsa, Jaakkola, M{\"a}kinen,
  Honkanen, Lindholm, Noponen, Ihalainen, Murtom{\"a}ki, Nojonen, Levo,
  Mertsalmi, Scheperjans, and
  Kaasinen]{nuuttilaDiagnosticAccuracyGlabellar2021}
Simo Nuuttila, Mikael Eklund, Juho Joutsa, Elina Jaakkola, Elina M{\"a}kinen,
  Emma~A. Honkanen, Kari Lindholm, Tommi Noponen, Toni Ihalainen, Kirsi
  Murtom{\"a}ki, Tanja Nojonen, Reeta Levo, Tuomas Mertsalmi, Filip
  Scheperjans, and Valtteri Kaasinen.
\newblock Diagnostic accuracy of glabellar tap sign for {{Parkinson}}'s
  disease.
\newblock \emph{Journal of Neural Transmission}, 128\penalty0 (11):\penalty0
  1655--1661, 2021.
\newblock ISSN 0300-9564.
\newblock \doi{10.1007/s00702-021-02391-3}.

\bibitem[VanderWerf et~al.(2003)VanderWerf, Brassinga, Reits, Aramideh, and
  {Ongerboer de Visser}]{vanderwerfEyelidMovementsBehavioral2003}
Frans VanderWerf, Peter Brassinga, Dik Reits, Majid Aramideh, and Bram
  {Ongerboer de Visser}.
\newblock Eyelid movements: Behavioral studies of blinking in humans under
  different stimulus conditions.
\newblock \emph{Journal of Neurophysiology}, 89\penalty0 (5):\penalty0
  2784--2796, May 2003.
\newblock ISSN 0022-3077.

\bibitem[Kwon et~al.(2013)Kwon, Shipley, Edirisinghe, Ezra, Rose, Best, and
  Cameron]{kwonHighspeedCameraCharacterization2013}
Kyung-Ah Kwon, Rebecca~J. Shipley, Mohan Edirisinghe, Daniel~G. Ezra, Geoff
  Rose, Serena~M. Best, and Ruth~E. Cameron.
\newblock High-speed camera characterization of voluntary eye blinking
  kinematics.
\newblock \emph{Journal of the Royal Society, Interface}, 10\penalty0
  (85):\penalty0 20130227, August 2013.
\newblock ISSN 1742-5662.
\newblock \doi{10.1098/rsif.2013.0227}.

\bibitem[Cruz et~al.(2011)Cruz, Garcia, Pinto, and
  Cechetti]{cruzSpontaneousEyeblinkActivity2011}
Antonio A.~V. Cruz, Denny~M. Garcia, Carolina~T. Pinto, and Sheila~P. Cechetti.
\newblock Spontaneous eyeblink activity.
\newblock \emph{The Ocular Surface}, 9\penalty0 (1):\penalty0 29--41, January
  2011.
\newblock ISSN 1542-0124.

\bibitem[Soukupova(2016)]{soukupovaRealTimeEyeBlink2016}
Tereza Soukupova.
\newblock Real-{{Time Eye Blink Detection}} using {{Facial Landmarks}}.
\newblock page~8, February 2016.
\newblock URL \url{https://api.semanticscholar.org/CorpusID:35923299}.

\bibitem[{Riverbank Computing Limited}(2023)]{pyqt6}
{Riverbank Computing Limited}.
\newblock Pyqt6, 2023.
\newblock URL \url{https://www.riverbankcomputing.com/software/pyqt/}.

\bibitem[Hochreiter et~al.(2023)Hochreiter, Hoche, Janik, Volk, Leistritz,
  Anders, and
  {Guntinas-Lichius}]{hochreiterMachineLearningBasedDetectingEyelid2023}
Jakob Hochreiter, Eric Hoche, Luisa Janik, Gerd~Fabian Volk, Lutz Leistritz,
  Christoph Anders, and Orlando {Guntinas-Lichius}.
\newblock Machine-{{Learning-Based Detecting}} of {{Eyelid Closure}} and
  {{Smiling Using Surface Electromyography}} of {{Auricular Muscles}} in
  {{Patients}} with {{Postparalytic Facial Synkinesis}}: {{A Feasibility
  Study}}.
\newblock \emph{Diagnostics}, 13\penalty0 (3):\penalty0 554, January 2023.
\newblock ISSN 2075-4418.
\newblock \doi{10.3390/diagnostics13030554}.

\bibitem[Chen et~al.(2021)Chen, Tzeng, Hsiao, Chen, Hung, and
  Lee]{chenSmartphoneBasedArtificialIntelligenceAssisted2021}
Hung-Chang Chen, Shin-Shi Tzeng, Yen-Chang Hsiao, Ruei-Feng Chen, Erh-Chien
  Hung, and Oscar~K. Lee.
\newblock Smartphone-{{Based Artificial Intelligence-Assisted Prediction}} for
  {{Eyelid Measurements}}: {{Algorithm Development}} and {{Observational
  Validation Study}}.
\newblock \emph{JMIR mHealth and uHealth}, 9\penalty0 (10):\penalty0 e32444,
  October 2021.
\newblock ISSN 2291-5222.
\newblock \doi{10.2196/32444}.

\bibitem[Lugaresi et~al.(2019)Lugaresi, Tang, Nash, McClanahan, Uboweja, Hays,
  Zhang, Chang, Yong, Lee, Chang, Hua, Georg, and
  Grundmann]{lugaresiMediaPipeFrameworkBuilding2019}
Camillo Lugaresi, Jiuqiang Tang, Hadon Nash, Chris McClanahan, Esha Uboweja,
  Michael Hays, Fan Zhang, Chuo-Ling Chang, Ming Yong, Juhyun Lee, Wan-Teh
  Chang, Wei Hua, Manfred Georg, and Matthias Grundmann.
\newblock {{MediaPipe}}: {{A}} framework for perceiving and processing reality.
\newblock In \emph{Third Workshop on Computer Vision for {{AR}}/{{VR}} at
  {{IEEE}} Computer Vision and Pattern Recognition ({{CVPR}}) 2019}, 2019.

\bibitem[Kartynnik et~al.(2019)Kartynnik, Ablavatski, Grishchenko, and
  Grundmann]{kartynnikRealtimeFacialSurface2019a}
Yury Kartynnik, Artsiom Ablavatski, Ivan Grishchenko, and Matthias Grundmann.
\newblock Real-time {{Facial Surface Geometry}} from {{Monocular Video}} on
  {{Mobile GPUs}}.
\newblock \emph{ArXiv}, abs/1907.06724, July 2019.
\newblock \doi{10.48550/arXiv.1907.06724}.

\bibitem[{The Qt Componany}(2023)]{qt6}
{The Qt Componany}.
\newblock Qt, 2023.
\newblock URL \url{https://www.qt.io/}.

\bibitem[Campagnola(2020)]{pyqtgraph}
Luke Campagnola.
\newblock Pyqtgraph: Scientific graphics and gui library for python, 2020.
\newblock URL \url{https://github.com/pyqtgraph/pyqtgraph}.

\bibitem[Virtanen et~al.(2020)Virtanen, Gommers, Oliphant, Haberland, Reddy,
  Cournapeau, Burovski, Peterson, Weckesser, Bright, {van der Walt}, Brett,
  Wilson, Millman, Mayorov, Nelson, Jones, Kern, Larson, Carey, Polat, Feng,
  Moore, VanderPlas, Laxalde, Perktold, Cimrman, Henriksen, Quintero, Harris,
  Archibald, Ribeiro, Pedregosa, {van Mulbregt}, {SciPy 1.0 Contributors},
  Vijaykumar, Bardelli, Rothberg, Hilboll, Kloeckner, Scopatz, Lee, Rokem,
  Woods, Fulton, Masson, H{\"a}ggstr{\"o}m, Fitzgerald, Nicholson, Hagen,
  Pasechnik, Olivetti, Martin, Wieser, Silva, Lenders, Wilhelm, Young, Price,
  Ingold, Allen, Lee, Audren, Probst, Dietrich, Silterra, Webber, Slavi{\v c},
  Nothman, Buchner, Kulick, Sch{\"o}nberger, {de Miranda Cardoso}, Reimer,
  Harrington, Rodr{\'i}guez, {Nunez-Iglesias}, Kuczynski, Tritz, Thoma,
  Newville, K{\"u}mmerer, Bolingbroke, Tartre, Pak, Smith, Nowaczyk, Shebanov,
  Pavlyk, Brodtkorb, Lee, McGibbon, Feldbauer, Lewis, Tygier, Sievert, Vigna,
  Peterson, More, Pudlik, Oshima, Pingel, Robitaille, Spura, Jones, Cera,
  Leslie, Zito, Krauss, Upadhyay, Halchenko, and
  {V{\'a}zquez-Baeza}]{virtanenSciPyFundamentalAlgorithms2020}
Pauli Virtanen, Ralf Gommers, Travis~E. Oliphant, Matt Haberland, Tyler Reddy,
  David Cournapeau, Evgeni Burovski, Pearu Peterson, Warren Weckesser, Jonathan
  Bright, St{\'e}fan~J. {van der Walt}, Matthew Brett, Joshua Wilson, K.~Jarrod
  Millman, Nikolay Mayorov, Andrew R.~J. Nelson, Eric Jones, Robert Kern, Eric
  Larson, C~J Carey, {\.I}lhan Polat, Yu~Feng, Eric~W. Moore, Jake VanderPlas,
  Denis Laxalde, Josef Perktold, Robert Cimrman, Ian Henriksen, E.~A. Quintero,
  Charles~R. Harris, Anne~M. Archibald, Ant{\^o}nio~H. Ribeiro, Fabian
  Pedregosa, Paul {van Mulbregt}, {SciPy 1.0 Contributors}, Aditya Vijaykumar,
  Alessandro~Pietro Bardelli, Alex Rothberg, Andreas Hilboll, Andreas
  Kloeckner, Anthony Scopatz, Antony Lee, Ariel Rokem, C.~Nathan Woods, Chad
  Fulton, Charles Masson, Christian H{\"a}ggstr{\"o}m, Clark Fitzgerald,
  David~A. Nicholson, David~R. Hagen, Dmitrii~V. Pasechnik, Emanuele Olivetti,
  Eric Martin, Eric Wieser, Fabrice Silva, Felix Lenders, Florian Wilhelm,
  G.~Young, Gavin~A. Price, Gert-Ludwig Ingold, Gregory~E. Allen, Gregory~R.
  Lee, Herv{\'e} Audren, Irvin Probst, J{\"o}rg~P. Dietrich, Jacob Silterra,
  James~T Webber, Janko Slavi{\v c}, Joel Nothman, Johannes Buchner, Johannes
  Kulick, Johannes~L. Sch{\"o}nberger, Jos{\'e}~Vin{\'i}cius {de Miranda
  Cardoso}, Joscha Reimer, Joseph Harrington, Juan Luis~Cano Rodr{\'i}guez,
  Juan {Nunez-Iglesias}, Justin Kuczynski, Kevin Tritz, Martin Thoma, Matthew
  Newville, Matthias K{\"u}mmerer, Maximilian Bolingbroke, Michael Tartre,
  Mikhail Pak, Nathaniel~J. Smith, Nikolai Nowaczyk, Nikolay Shebanov,
  Oleksandr Pavlyk, Per~A. Brodtkorb, Perry Lee, Robert~T. McGibbon, Roman
  Feldbauer, Sam Lewis, Sam Tygier, Scott Sievert, Sebastiano Vigna, Stefan
  Peterson, Surhud More, Tadeusz Pudlik, Takuya Oshima, Thomas~J. Pingel,
  Thomas~P. Robitaille, Thomas Spura, Thouis~R. Jones, Tim Cera, Tim Leslie,
  Tiziano Zito, Tom Krauss, Utkarsh Upadhyay, Yaroslav~O. Halchenko, and
  Yoshiki {V{\'a}zquez-Baeza}.
\newblock {{SciPy}} 1.0: Fundamental algorithms for scientific computing in
  {{Python}}.
\newblock \emph{Nature Methods}, 17\penalty0 (3):\penalty0 261--272, March
  2020.
\newblock ISSN 1548-7091, 1548-7105.
\newblock \doi{10.1038/s41592-019-0686-2}.

\bibitem[Otsu(1979)]{otsu}
Nobuyuki Otsu.
\newblock A threshold selection method from gray-level histograms.
\newblock \emph{IEEE Transactions on Systems, Man, and Cybernetics}, 9\penalty0
  (1):\penalty0 62--66, 1979.
\newblock \doi{10.1109/TSMC.1979.4310076}.

\bibitem[ana(2020)]{anaconda}
Anaconda software distribution, 2020.
\newblock URL \url{https://docs.anaconda.com/}.

\end{thebibliography}

\end{document}